\pdfoutput=1

\documentclass[11pt]{article}

\usepackage[final]{coling}

\usepackage{times}
\usepackage{latexsym}

\usepackage[T1]{fontenc}

\usepackage[utf8]{inputenc}

\usepackage{microtype}

\usepackage{inconsolata}

\usepackage{graphicx}
\usepackage{xcolor}
\usepackage{booktabs}
\usepackage{algorithm}
\usepackage{algpseudocode}
\usepackage{amsmath}
\usepackage{multirow}
\usepackage{enumitem}
\usepackage{fontawesome5}
\usepackage{hyperref}
\usepackage{fancyvrb}
\usepackage{float}
\usepackage{alltt}

\title{Automated Clinical Data Extraction with Knowledge Conditioned LLMs}

\author{
  Diya Li, Asim Kadav, Aijing Gao\textsuperscript{\dag}, Rui Li, Richard Bourgon\textsuperscript{\dag} \\
  \textsuperscript{\dag}Freenome, South San Francisco, CA, USA \\
  \textsuperscript{\dag}\texttt{\{aijing.gao, richard.bourgon\}@freenome.com} \\
  \texttt{\{916lidiya, asimkadav, raylee8023\}@gmail.com}
}  

\begin{document}
\maketitle
\begin{abstract}
The extraction of lung lesion information from clinical and medical imaging reports is crucial for research on and clinical care of lung-related diseases. 
Large language models (LLMs) can be effective at interpreting unstructured text in reports, but they often hallucinate due to a lack of domain-specific knowledge, leading to reduced accuracy and posing challenges for use in clinical settings. 
To address this, we propose a novel framework that aligns generated internal knowledge with external knowledge through in-context learning (ICL). 
Our framework employs a retriever to identify relevant units of internal or external knowledge and a grader to evaluate the truthfulness and helpfulness of the retrieved internal-knowledge rules, to align and update the knowledge bases.
Experiments with expert-curated test datasets demonstrate that this ICL approach can increase the F1 score for key fields (lesion size, margin and solidity) by an average of 12.9\% over existing ICL methods.
\end{abstract}
\section{Introduction}
\label{sec:introduction}

\begin{figure}[ht]
\centering
\includegraphics[width=0.5\textwidth]{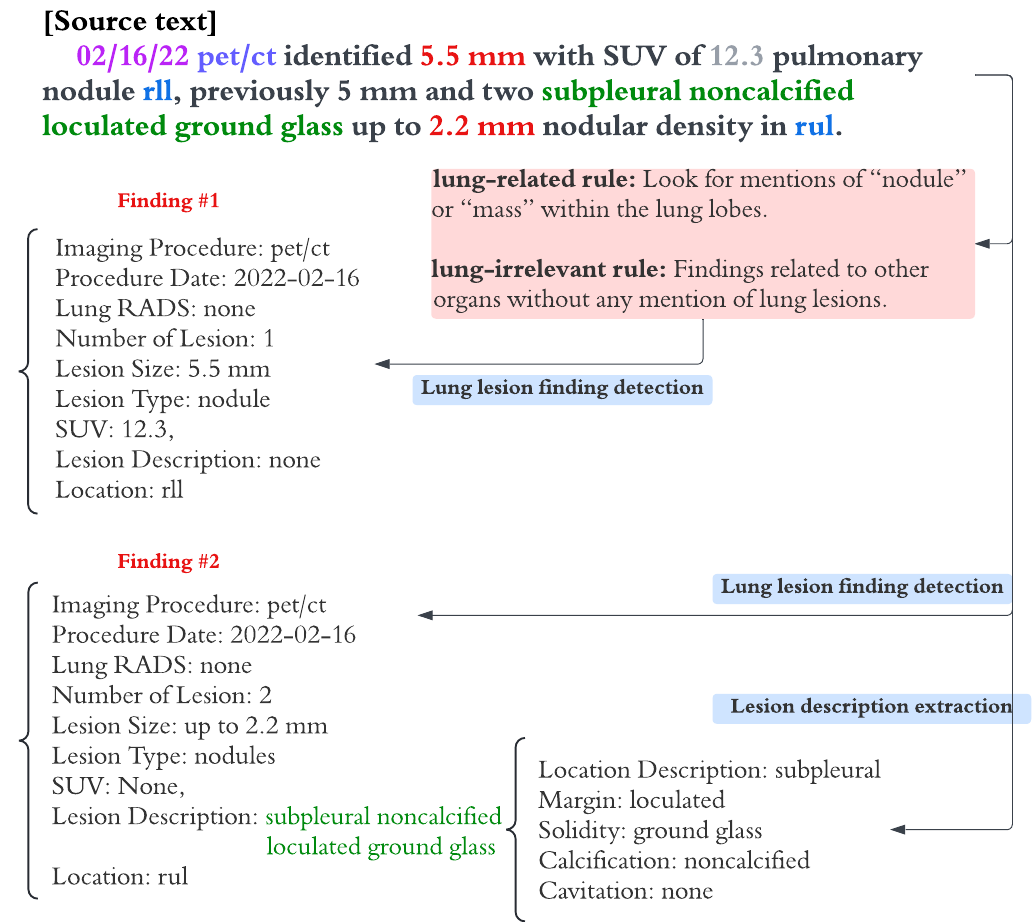}
\caption{Example of lung lesion information extraction. Two findings (one describing a single lesion, and the other, two lesions) were identified in the source text. Example rules from the generated internal knowledge base are shown. First-stage finding detection and primary structured field parsing is followed by a second stage that further parses lesion description text.}
\label{fig:example}
\end{figure}

Lung lesion clinical data extraction from medical imaging and clinical reports plays a crucial role in enhancing the early detection and study of lung-related diseases~\citep{zhang2018pulmonary, huang2024critical}. 
Accurate automated extraction can reduce the manual effort required by a radiologist or physician. 
As illustrated in Figure \ref{fig:example}, given a report, the task is to automatically extract information at the \textit{finding} level, where a \textit{finding} refers to text describing one or more closely related lesions.
Since a report can have multiple findings, our task is to detect all findings and to parse each of them into a structured schema with pre-defined fields. (See Figure \ref{fig:example} and Tables \ref{tab:finding_detection} and \ref{tab:char_extraction}).

However, interpreting unstructured text in reports presents a considerable challenge due to the complexity and variability of medical language \cite{wang2018clinical}.
Creating specialized supervised machine learning models for specific medical terms can be effective but is often resource-intensive~\cite{spasic2020clinical}. 
Recently, large language models (LLMs) have emerged as valuable assistive tools for general clinical data extraction \cite{singhal2023medpalm,thirunavukarasu2023llmInMed}.

Nonetheless, using LLMs for clinical data extraction suffers from several challenges. 
First, LLMs often miss fine-grained details in clinical data extraction~\cite{ji2023survey, dagdelen2024structured}, due to a lack of domain-specific knowledge. 
The extraction of lung lesion information requires an understanding of specialized fields (such as \textit{margin} and \textit{solidity}) that are not included in predefined schemas~\cite{LDC2006,Consortium2008}.
Second, for extracting complex domain-specific fields, LLMs often fail to understand nested subfields \cite{chen2024ee}, and as a result, they may generate structurally inconsistent outputs.

To provide an automated method of clinical data extraction that addresses the above limitations, we propose a two-stage LLM framework that uses an {\em internal knowledge base} that is iteratively aligned with an expert-derived {\em external knowledge base} using in-context learning (ICL). 
Specifically, we first create the internal knowledge base by utilizing a manually curated medical report training corpus to generate {\em references}.
The references that are deemed relevant to new input reports are converted into a set of higher-level {\em rules} that comprise the internal knowledge base. 
When extracting data from a report, rules from the internal knowledge base are {\em retrieved} and {\em graded} by our system to improve alignment with the external knowledge base.
This process enhances the effectiveness of finding detection by leveraging relevant extraction patterns that are aligned with external knowledge.
Lastly, to address the challenge of extraction of nested fields, we first extract an unstructured lesion description text field for each finding, then parse the description text into structured fields as a separate task that employs a more instructed approach (Figure \ref{fig:example}).

We validate our approach through experiments using a curated dataset from a real-world clinical trial that includes annotations from medical experts.  
In addition, we define a new field schema for the lung lesion extraction task that may be useful for related lung disease studies.
Our results demonstrate improvements in the accuracy of lung lesion clinical data extractions when using our framework compared to existing ICL methods. 

\section{Methodology}
\label{sec:methodology}

\begin{figure*}[h]
\centering
\includegraphics[width=\textwidth]{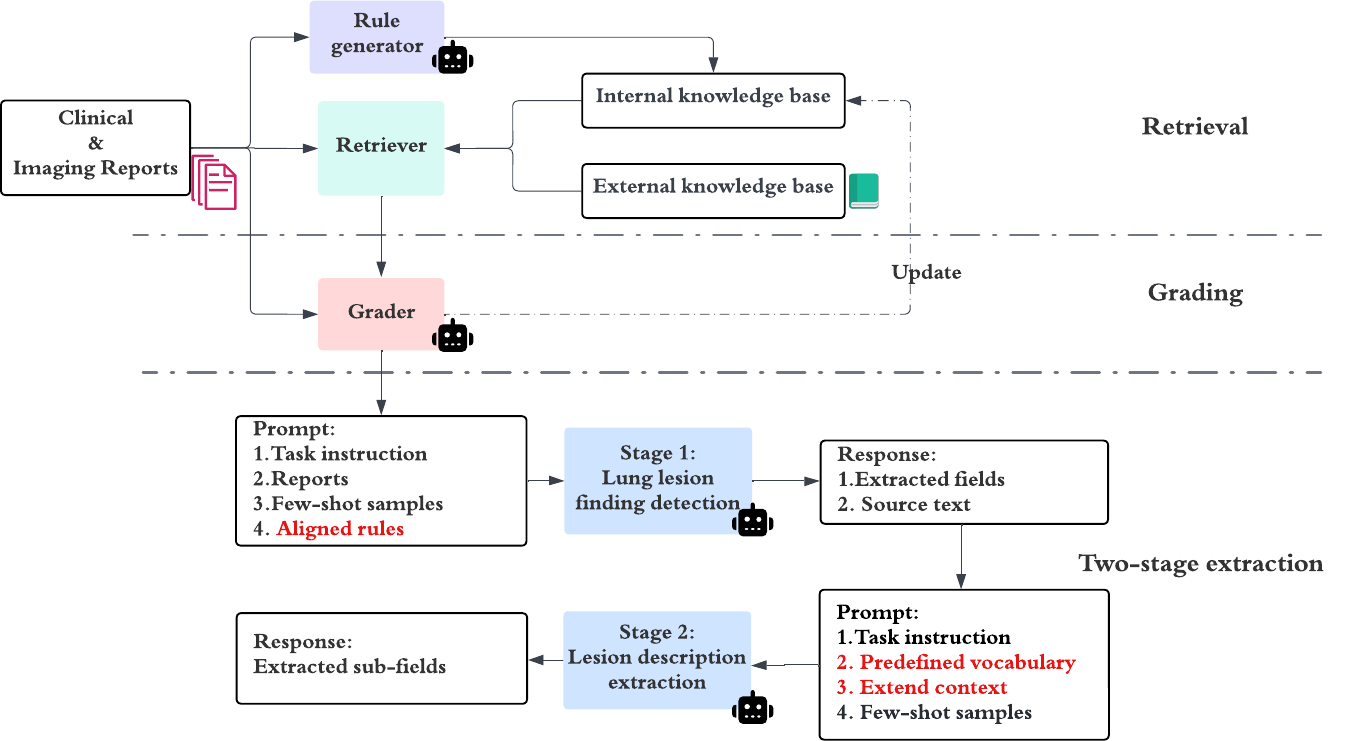}
\caption{Framework for two-stage knowledge conditioned clinical data extraction. 
The \faRobot~symbol indicates that the module is implemented by prompting an LLM. Rules used in prompts for lesion finding detection are derived from the internal knowledge base and aligned with external knowledge by a grader. 
Unstructured lesion description text is extracted in stage 1. In stage 2, this text is parsed into structured fields by providing the LLM with additional specialized inputs, including a controlled vocabulary.}
\label{fig:framework}
\end{figure*}


\subsection{Task Definition}
Our task is to extract lung lesions findings from clinical and imaging reports.
Key fields include imaging procedure, lesion size, margin, solidity, lobe, and for PET/CT, standardized uptake value (SUV). \footnote{Details on the meaning of these fields, along with a complete list of extraction fields, are provided in Table \ref{tab:guidelines}.}
Extraction of the above fields is useful for oncology research and to support clinical care~\cite{LungCancerStaging}.

\subsection{Clinical Data Extraction Framework}
Given input reports $\mathcal{X}$, an internal knowledge base (KB) containing LLM-generated rules $\mathcal{D} = \{d_1, ..., d_N\}$, and an expert-curated external KB $\mathcal{K} = \{k_1, ..., k_M\}$, our system aims to generate the extracted fields as $\mathcal{Y}$.

Our framework for aligning the KBs uses a retriever $\mathcal{R}$ and a grader $\mathcal{G}$. 
The retriever $\mathcal{R}$ retrieves the top $k$ rules $\tilde{\mathcal{D}} = \{d_{1}, ..., d_{k}\}$ that are relevant to the input $\mathcal{X}$ from $\mathcal{D}$. 
The grader $\mathcal{G}$ then further selects and attempts to improve the rules $\tilde{\mathcal{D}}$ based on the input $\mathcal{X}$ and retrieved external knowledge $\tilde{\mathcal{K}} = \mathcal{R}(\mathcal{K} | \tilde{\mathcal{D}})$ from $\mathcal{K}$, resulting in $\hat{\mathcal{D}} = \mathcal{G}(\tilde{\mathcal{D}}| \mathcal{X}, \tilde{\mathcal{K}})$, where $\hat{\mathcal{D}}$ are knowledge aligned rules. 
By adding the improved $\hat{\mathcal{D}}$ to the default prompt, an LLM extracts the fields from reports $\mathcal{X}$ into our structured lesion field schema $\mathcal{Y}$.

\subsection{Lung Lesion Knowledge Base Construction}
We construct two KBs: an internal one generated by an LLM-based rule generator module using a small labeled training set, and an external one using expert knowledge resources.

\paragraph{Internal Knowledge Base Construction}
Using a small training set of annotated reports with lung lesion and non-lung lesion findings, we first ask the \emph{rule generator} (implemented by an LLM) to create \emph{lung-related} and \emph{lung-irrelevant references}.
A reference takes the following form:

\begin{description}
    \item[source text] ``Additional soft tissue nodular density in the right upper lobe measuring 1.3 cm.''
    \item[explanation] ``This finding is described as a 'soft tissue nodular density' measuring 1.3 cm, located in the right upper lobe. It may indicate a small mass or nodule.''
\end{description}

To transform these references into more general, reusable \emph{rules}, we next prompt the rule generator to identify common properties among the references and to extrapolate.
For example, lung-related references often include measurements, so an extrapolated rule might be: ``\textit{Look for descriptions that include measurements (e.g., `identified 5.5 mm', `measures 1.8 x 1.2 cm') which often indicate lung lesions}.'' 
The rules are generated in a multi-dialogue style, and the generation process is illustrated in Table \ref{tab:rule_generator}.
These generalized rules make up our internal knowledge base, denoted as $\mathcal{D}$, consisting of \emph{lung-related} and \emph{lung-irrelevant rules}.
Example rules are provided in Table~\ref{tab:rule}. 
We prioritize the \emph{lung-irrelevant rules} since they assist LLMs in distinguishing findings that are not related to the lungs, thereby reducing false positives in inference.

\paragraph{External Knowledge-base Construction}
We manually identify external domain-specific knowledge from authoritative sources, including clinical guidelines\footnote{https://radiopaedia.org/}\textsuperscript{,}\footnote{https://radiologyassistant.nl/}\textsuperscript{,}\footnote{https://www.cancer.org/cancer/types/lung-cancer/detection-diagnosis-staging.html} and expert medical opinions. 
The collected information is divided into chunks and stored in a local database for easy retrieval.
The chunk size is 1000 with an overlap of 200 for indexing.
The external KB, denoted as $\mathcal{K}$, encompasses a diverse range of content formats, including structured data, textual information, and procedural guidelines.

\subsection{Retriever \& Grader}
\paragraph{Retriever}
Given reports $\mathcal{X}$, the retriever module $\mathcal{R}$ is responsible for identifying the top-$k$ relevant \textit{lung-related} and \textit{lung-irrelevant rules} from the internal KB $\mathcal{D}$. 
This retrieval process matches the input reports with the most pertinent rules and returns these as $\tilde{\mathcal{D}} = \mathcal{R}(\mathcal{D} | \mathcal{X})$. 

For each rule $d \in \tilde{\mathcal{D}}$, the retriever $\mathcal{R}$ also retrieves $\tilde{\mathcal{K}} = \mathcal{R}(d, \mathcal{K})$ from the external KB $\mathcal{K}$ for use by the grader $G$ for knowledge alignment.

\begin{algorithm}[th]
\caption{Grading Algorithm}
\label{algo:grading}
\textbf{Input}: imaging and clinical reports $\mathcal{X}$, retriever $\mathcal{R}$, grader $\mathcal{G}$, retrieved internal rules $\tilde{\mathcal{D}}$, external knowledge $\mathcal{K}$, number of iterations $I$, thresholds. \\
\textbf{Output}: aligned rules $\hat{\mathcal{D}}$, updated internal KB $\mathcal{D}$.

\begin{algorithmic}[1]
\State Initialize $\hat{\mathcal{D}} = \emptyset$;
\For{$i = 1$ to $I$}
    \For{$d \in \tilde{\mathcal{D}}$}
        \State $\tilde{\mathcal{K}} = \mathcal{R}(d, \mathcal{K})$;
        \State $T = \mathcal{G}_\text{truthfulness}(d, \tilde{\mathcal{K}})$;
        \If{$T < \text{threshold}_T$}
            \State $\mathcal{D} = \mathcal{D} \setminus \{d\}$;
            \State $d = \mathcal{G}_\text{align}(d, \tilde{\mathcal{K}})$;
            \State $\mathcal{D} = \mathcal{D} \cup \{d\}$;
        \EndIf
        \State $\hat{\mathcal{D}} = \hat{\mathcal{D}} \cup \{d\}$;
    \EndFor
    \For{$d \in \hat{\mathcal{D}}$}
        \State $H = \mathcal{G}_\text{helpfulness}(d, \mathcal{X})$;
        \If{$H < \text{threshold}_H$}
            \State $\hat{\mathcal{D}} = \hat{\mathcal{D}} \setminus \{d\}$;
        \EndIf
    \EndFor
\EndFor
\end{algorithmic}
\end{algorithm}

\paragraph{Grader}
To improve the quality of the retrieved rules $\tilde{\mathcal{D}}$, we introduce a grader $\mathcal{G}$, also implemented with an LLM. 
The grader is assigned two tasks and iterates over these tasks to refine the rules in internal KB $\mathcal{D}$.

First, $\mathcal{G}$ grades the rules in $\tilde{\mathcal{D}}$ with a \textit{truthfulness} score, an integer ranging from 1 to 3, by comparing each $d$ against retrieved external knowledge $\tilde{\mathcal{K}} = \mathcal{R}(d, \mathcal{K})$ and assessing its alignment with authoritative sources. 
If the \textit{truthfulness} score of a rule falls below a threshold, the grader removes the rule from $\mathcal{D}$ and generates revised rules that are added back to $\mathcal{D}$. 
Second, the grader $\mathcal{G}$ assigns the aligned rules in $\hat{\mathcal{D}}$ a \textit{helpfulness} score based on their relevance to the input reports $\mathcal{X}$. 
The \textit{helpfulness} score is an integer ranging from 1 to 5.
To assess \textit{helpfulness}, the grader analyzes how well each rule supports the extraction and interpretation of information from $\mathcal{X}$.
Rules that do not meet the helpfulness threshold are removed from $\hat{\mathcal{D}}$.
This process is repeated for each rule $d$ over $I$ iterations, with $I$ determined through practical experience.
This iterative approach helps refine the alignment of the rules, ensuring that only the most relevant ones are retained in $\hat{\mathcal{D}}$.
The prompts for assessing \textit{truthfulness} and \textit{helpfulness} can be found in Table \ref{tab:grader_prompt}. 

The final set of high-scoring rules $\hat{\mathcal{D}}$ is used in prompts for lesion finding extractions. This iterative process is intended to increase the likelihood that rules in the updated $\hat{\mathcal{D}}$ yield outputs with the desired properties. The full grading algorithm is detailed in Algorithm \ref{algo:grading}, where $\mathcal{G}_\text{align}$ returns the aligned rule based on retrieved external knowledge.

\subsection{Two-stage Extraction}
Clinical data often contain nested information. For example, an imaging report may include two findings described in a single phrase: \textit{``2 adjacent pulmonary nodules within the left lower lobe, the larger of the two measuring 5mm with an SUV of 2.39.''} 
In cases like this, the LLM often fails to detect the second finding because it is not well separated from the first finding in the text.

To address this limitation, we decompose the clinical data extraction task into two stages: (i) lung lesion finding detection and primary structured field parsing, followed by (ii) further parsing of lesion description text.

For the first stage, we use $\hat{\mathcal{D}}$ as a part of the LLM prompt for the lung lesion finding detection, along with task instructions, the input reports, and few-shot samples (Table \ref{tab:finding_detection}). 
The second stage aims to extract additional structured fields from the \textit{lesion description} text.
$\hat{\mathcal{D}}$ does not contribute in the second stage, as the set of valid terms to describe lesion description fields is limited.
Instead, we provide the LLM with a controlled vocabulary based on the SNOMED ontology~\cite{SNOMED} (Table \ref{tab:char_extraction}). 
We also note that the \textit{lesion description} alone is often insufficient, since information missed in the first stage can lead to errors in subsequent extraction steps. 
To mitigate this issue, the second stage prompt combines the extracted \textit{lesion description} text with the full \textit{source text} from the first stage, to extend the context for extracting lesion description fields.

The two-stage extraction workflow is illustrated in Figure \ref{fig:framework}.
\section{Experiments}
\label{sec:experiments}

\begin{table*}[ht]
\centering
\small
\begin{tabular}{c|l|ccc|ccc|ccc|ccc}
\toprule
\multirow{2}{*}{\textbf{Stage}}& \multirow{2}{*}{\textbf{Fields}} & \multicolumn{3}{c}{\textbf{Default Prompts}} & \multicolumn{3}{c}{\textbf{CoT}} & \multicolumn{3}{c}{\textbf{RAG}} & \multicolumn{3}{c} {\textbf{Ours}} \\
\cmidrule(lr){3-5} \cmidrule(lr){6-8} \cmidrule(lr){9-11} \cmidrule(lr){12-14}
& & \textbf{P} & \textbf{R} & \textbf{F1} & \textbf{P} & \textbf{R} & \textbf{F1} & \textbf{P} & \textbf{R} & \textbf{F1} & \textbf{P} & \textbf{R} & \textbf{F1} \\
\midrule
\multirow{5}{*}{\# 1} & Image procedure & 78.2 & 63.6 & 70.1 & 79.5 & 64.2 & \underline{71.0} & 75.3 & 61.1 & 67.5 & 86.5 & 72.7 & \textbf{79.0} \\
& Lesion size \dag & 85.9 & 82.1 & 84.0 & 87.3 & 84.2 & \underline{85.7} & 83.0 & 78.5 & 80.7 & 92.8 & 85.8 & \textbf{89.1} \\
& SUV & 76.0 & 73.1 & 74.5 & 77.4 & 74.3 & \underline{75.8} & 72.5 & 69.4 & 70.9 & 86.6 & 74.0 & \textbf{79.7} \\
& Lesion type & 83.7 & 67.3 & 74.6 & 85.1 & 68.7 & \underline{76.1} & 80.2 & 63.9 & 71.2 & 88.1 & 73.6 & \textbf{80.2} \\
& Lobe & 72.7 & 60.4 & 66.0 & 74.0 & 61.5 & \underline{67.2} & 70.0 & 57.5 & 63.0 & 81.9 & 69.6 & \textbf{75.2} \\
\midrule
\multirow{4}{*}{\#2} & Margin \dag & 68.4 & 65.0 & 66.7 & 68.5 & 67.5 & 67.5 & 75.0 & 63.2 & \underline{68.6} & 90.0 & 76.3 & \textbf{82.4} \\
& Solidity \dag & 65.0 & 35.7 & 45.7 & 67.3 & 36.9 & \underline{47.7} & 77.1 & 27.8 & 40.7 & 96.9 & 55.6 & \textbf{69.2} \\
& Calcification & 87.7 & 61.0 & 71.6 & 89.0 & 62.1 & \underline{73.2} & 76.0 & 62.8 & 67.5 & 88.8 & 67.8 & \textbf{75.7} \\
& Cavitation & 50.0 & 100.0 & 66.7 & 60.0 & 100.0 & \underline{73.4} & 50.0 & 100.0 & 66.7 & 87.5 & 100.0 & \textbf{91.7 }\\
\bottomrule
\end{tabular}
\caption{Overall precision (P), recall (R), and F1 scores evaluated on the test set. 
The results are averaged over 5 runs. 
The best results are marked in bold, and second best are underlined. 
Fields extracted in stage 1 vs. stage 2 are indicated. 
Fields marked with \(\dag\) are the clinically most important fields.}
\label{tab:results}
\end{table*}

\subsection{Datasets}
Our work utilizes clinical and imaging reports from Freenome's Vallania study (ClinicalTrials.gov, NCT05254834), which include lung cancer screening and other clinical results.
Clinical experts manually identify all lung lesion findings and extract relevant fields based on our annotation schema (Table \ref{tab:guidelines}). 
To develop a gold standard dataset for performance evaluation, 19 subjects are randomly sampled.
These subjects have a total of 31 clinical and 30 imaging reports, resulting in 189 findings.
We randomly select 9 of these subjects as the training set, with the remaining 10 designated as the test set.
Dataset and annotation details are discussed in Appendix~\ref{sec:appendix_data}.

\subsection{Evaluation Metrics}
For a given test report, the gold standard findings and the system-detected findings may differ in number and/or ordering. 
The two sets of findings need to be aligned to one other.
To achieve this, we perform an additional matching step and use the Hungarian algorithm\footnote{\url{https://en.wikipedia.org/wiki/Hungarian_algorithm}} to match the gold-standard and system-detected findings. 
All extracted fields are used to construct the cost matrix for matching.
We report micro precision, recall, and F1 scores for the extraction task.

\subsection{Module Implementation}
The LLMs used for rule generation, the grader, lung lesion finding detection, and lesion description extraction are based on the official API of the Google PaLM2 model~\cite{anil2023palm}.
All prompts used with LLMs are listed in Appendix \ref{appendix:prompt}.

We use retriever $\mathcal{R}$ to obtain the top $k$ relevant internal knowledge rules $\tilde{\mathcal{D}} = \mathcal{R}(\mathcal{D} | \mathcal{X})$ and retrieve external knowledge $\tilde{\mathcal{K}} = \mathcal{R}(\tilde{\mathcal{D}}, \mathcal{K})$ based on semantic similarity to $\tilde{\mathcal{D}}$. 
Specifically, we use the text embedding API (\textit{text-embedding-004}) from Google~\cite{GoogleCloudVertexAI} to obtain the embeddings of $\mathcal{X}$, $\mathcal{D}$, and $\mathcal{K}$. 
Cosine similarity is used for semantic similarity scores.
For hyper-parameter settings used in our system, refer to Table~\ref{tab:parameter}.

\subsection{Comparison Baselines}
As there is no prior work on lung lesion extraction using LLMs with our curated real-world dataset, we apply commonly-used ICL baseline methods and compare against the following:

\paragraph{Few-shot Learning}
Here, the LLM is provided with a small number of gold standard examples as a part of the basic prompts used for lesion finding detection and lesion description extraction \cite{brown22fewshot}. 
These prompts, referred to as \textit{default prompts}, do not include any knowledge base content or additional guidance. We report results based on these default prompts, and other methods incrementally build upon them.

\paragraph{Chain of Thought (CoT)}
Additional instructions are added in the default prompts to guide the LLM to break down the lesion finding detection task into simpler, sequential steps by \textit{thinking step by step} \cite{wei2022chain}. 
CoT is not applied at stage-two because this task is straightforward to conduct.

\paragraph{Retrieval Augmented Generation (RAG)}
RAG complements basic LLM queries, and it attempts to reduce hallucination by introducing external knowledge to improve the context~\cite{lewis2020retrieval}.
We implement a RAG approach that directly retrieves information from $\mathcal{K}$ and adds the retrieved external knowledge chunks into the default prompt. This approach does not use the internal KB ($\mathcal{D}$).

\begin{table*}[ht]
\centering
\small
\begin{tabular}{c|l|ccc|ccc|ccc|ccc}
\toprule
\multirow{2}{*}{\textbf{Stage}} & \multirow{2}{*}{\textbf{Fields}} & \multicolumn{3}{c}{\textbf{w/o knowledge}} & \multicolumn{3}{c}{\textbf{w/o context}} & \multicolumn{3}{c}{\textbf{w/o grading}} & \multicolumn{3}{c} {\textbf{Ours}} \\
\cmidrule(lr){3-5} \cmidrule(lr){6-8} \cmidrule(lr){9-11} \cmidrule(lr){12-14}
& & \textbf{P} & \textbf{R} & \textbf{F1} & \textbf{P} & \textbf{R} & \textbf{F1} & \textbf{P} & \textbf{R} & \textbf{F1} & \textbf{P} & \textbf{R} & \textbf{F1} \\
\midrule
\multirow{5}{*}{\# 1} & Image procedure & 78.2 & 63.6 & 70.1 & 86.5 & 72.7 & \textbf{79.0} & 84.6 & 61.7 & 71.4 & 86.5 & 72.7 & \textbf{79.0} \\ 
& Lesion size \dag & 85.9 & 82.1 & 84.0 & 92.8 & 85.8 & \textbf{89.1} & 92.4 & 85.1 & 88.5 & 92.8 & 85.8 & \textbf{89.1}\\
& SUV & 76.0 & 73.1 & 74.5 &  86.6 & 74.0 & \textbf{79.7} & 85.7 & 69.2 & 76.6 & 86.6 & 74.0 & \textbf{79.7} \\
& Lesion type  & 83.7 & 67.3 & 74.6 & 88.1 & 73.6 & \textbf{80.2} & 91.2 & 69.8 & 78.9 & 88.1 & 73.6 & \textbf{80.2} \\
& Lobe & 72.7 & 60.4 & 66.0 & 81.9 & 69.6 & \textbf{75.2} & 80.8 & 63.2 & 70.8 & 81.9 & 69.6 & \textbf{75.2} \\
\midrule
\multirow{4}{*}{\# 2}  & Margin \dag & 68.6 & 67.8 & 67.2 & 85.8 & 75.0 & 80.0 & 84.5 & 66.7 & 74.1 & 90.0 & 76.3 & \textbf{82.4} \\
& Solidity \dag & 91.7 & 37.0 & 52.6 & 95.0 & 44.4 & 60.1 & 90.0 & 55.7 & 65.8 & 96.9 & 55.6 & \textbf{69.2} \\
& Calcification & 100.0 & 57.1 & 72.7 & 80.4 & 64.3 & 70.1 & 83.3 & 71.4 & 73.3 & 88.8 & 67.8 & \textbf{75.7} \\
& Cavitation & 55.6 & 100.0 & 71.5 & 75.0 & 100.0 & 83.4 & 66.7 & 100.0 & 77.8 & 87.5 & 100.0 & \textbf{91.7} \\
\bottomrule
\end{tabular}
\caption{Ablation study on our two-stage knowledge conditioned model.
Note that the performance of the model in the first stage without extended context (``w/o context'') is the same as our full model, as the context extension is only applied during the second extraction stage.}
\label{tab:ablation}
\end{table*}

\subsection{Results and Analysis}
\paragraph{Overall results}
The overall results are shown in Table \ref{tab:results}. 
We are especially interested in the fields denoted with \(\dag\), which include \textit{lesion size}, \textit{margin}, and \textit{solidity}, because these are often of greatest clinical interest for cancer work~\cite{nathan1962differentiation, khan2011solitary}.

In our experiments, the benefit of Chain of Thought (CoT) reasoning is limited, as it appears to be more effective for traditional multi-step reasoning tasks, rather than our specialized extraction task~\cite{NEURIPS2022_9d560961}. 
The RAG implementation also performs poorly in the lung lesion extraction task --- even worse than the default prompts. 
This may be due to incorrect retrieval of external knowledge based only on semantic similarity search, resulting in adding noise to the prompt. 
This suggests that the utility of the external knowledge ($\mathcal{K}$) may be constrained without first attempting to align it to the specific extraction task.
Unlike RAG, our method first generates internal knowledge related to the specific extraction task. 
External knowledge is then utilized solely to align and update the internal knowledge. 
Results in Table \ref{tab:results} suggest that this improves the quality of our method's generated rules.

Our model outperforms all ICL baselines across all fields, particularly excelling in the \(\dag\) fields, with an average of 12.9\% increase in F1 score. 
Specifically, it achieves a 3.4\% improvement in \textit{lesion size}, over 13.8\% in \textit{margin}, and a 21.5\% improvement in \textit{solidity}.

\paragraph{Ablation Study}
To assess the contribution of each component of our method, we conduct ablation tests by removing each main module. 
The ablation results are listed in Table \ref{tab:ablation}. 

Notably, there is a significant performance decrease in the model that does not use the knowledge bases (``w/o knowledge''), indicating the importance of incorporating domain knowledge. Further, because lesion finding extraction quality degrades when the KBs are ignored, the quality of stage 2 lesion description extraction also degrades.
Next, the model that omits providing extended context and the SNOMED controlled vocabulary for stage 2 (``w/o context''), performs worse for stage 2 fields. 
This indicates that extended context in stage 2 prompts can help prevent error propagation from stage 1, and that the controlled vocabulary standardizes the extraction of lesion description fields.
Finally, we observe that the performance of the model that does not use the grader for knowledge alignment (``w/o grading'') varies significantly across runs, suggesting that the grader's alignment role improves consistency and reduces noise.

\subsection{Discussion}
\begin{table*}[ht]
    \centering
    \small
    \begin{tabular}{l|c|p{12cm}}
    \toprule
    \textbf{Category} & \textbf{Rule \#} &\textbf{Picked Rule} \\
    \midrule
    \multirow{3}{*}{Lung-related} & \#1 & \{
        "pattern": "solid | partly solid | groundglass",
        "rule": "Clinical notes mentioning `solid', `partly solid', `groundglass' could indicate pulmonary nodule findings."
    \} \\
        & \#2 & \{
                "pattern": "nodule",
                "rule": "Look for descriptions that mention `nodule', which often indicate lung lesions."
            \} \\
            
        & \#3 &  \{
                "pattern": "mass",
                "rule": "Look for descriptions that mention `mass' which often indicate lung lesions."
            \} \\
    \midrule
    Lung-irrelevant & \#4 & \{
            "pattern": "liver | kidney | other organs",
            "rule": "Findings related to other organs (e.g., liver, kidney) without any mention of lung lesions."
          \}\\
    
    \bottomrule
    \end{tabular}
    \caption{Most frequently picked lung-related and lung-irrelevant rules in test dataset.}
    \label{tab:rule}
    \end{table*}
    

\paragraph{Case Study of Internal Knowledge}
In our knowledge conditioned model, the grader iteratively updates the internal knowledge if a rule's \textit{truthfulness} score falls below a threshold, which is a hyper-parameter in our experiments. 

To better understand the impact of the aligned rules in the internal KB, we identify the most frequently picked lung-related and lung-irrelevant rules from the test set (Table \ref{tab:rule}). 
Rules about \textit{nodules} and \textit{masses} are frequently picked, as these are two commonly used terms for lung lesion types. (See rules \#2 and \#3 in Table \ref{tab:rule}.)
We also observe that the LLM tends be better at detecting lung lesion findings with explicit lesion sizes, using these as an anchor point to extract the full finding. 
\textit{Solidity} information is sparse in clinical data, but there are many cases where the finding does not have size information yet it describes solidity. 
Terms like \textit{solid}, \textit{partly solid}, and \textit{groundglass} often refer to the solidity field. 
Rule \#1 in Table \ref{tab:rule} contributes to the LLM's ability to detect lesion findings that reference lesion solidity.

For lung-irrelevant rules, the top-picked rule relates to findings in other organs, such as liver and kidney, without any mention of lung. 
Obviously, a rule of this type helps in distinguishing between relevant and irrelevant findings.

\paragraph{Effect of Retriever Top-$k$}
\begin{figure}[h]
\centering
\includegraphics[width=0.5\textwidth]{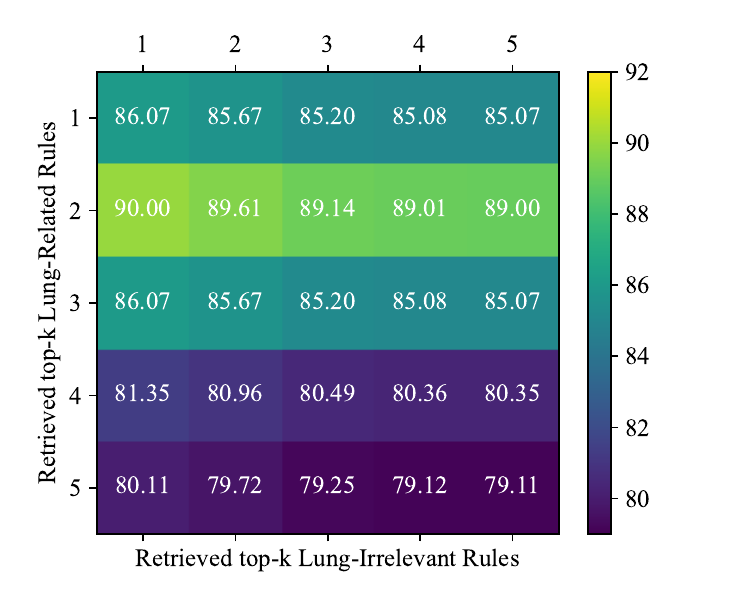}
\caption{Heatmap of lesion size extraction performance with varying values for the retriever's top-$k$ hyper-parameter, for both lung-related and lung-irrelevant rules.}
\label{fig:top_k}
\end{figure}

To determine the optimal $k$ values for internal knowledge retrieval, we perform a grid search using the training set, evaluating the performance of \textit{lesion size} extraction. Different values for $k$ are considered for both lung-related and lung-irrelevant rules. 
As shown in Figure \ref{fig:top_k}, the best extraction performance was observed when $k=2$ for lung-related rules and $k=1$ for lung-irrelevant rules. 
We use these optimal $k$ values for extraction in the test set. 
An interesting finding is that using only a few rules contributes significantly to improving lesion size extraction performance.
\section{Related Work}
\label{sec:relatedwork}

\subsection{Clinical Information Extraction}
Early work in clinical information extraction focused on rule-based systems and supervised machine learning techniques~\cite{savova2010discovering, wang2018clinical, barrett2013engineering, denny2010phewas, mehrabi2015identification, roberts2012machine, li2015end}, which were labor-intensive to create and required a process that lacked scalability. 

Recently, deep learning models, especially transformer-based architectures, have shown promise in clinical information extraction~\cite{zhong2022improving, spasic2020clinical}. 
These models reduce the need for extensive feature engineering, but they rely on high-quality annotated data. 

In the past few years, LLMs have been applied to clinical information extraction~\cite{goel2023llms, wornow2024zeroshot}. 
LLMs can extract multiple fields simultaneously without requiring labeled training data for each field.
While LLMs show promise in accelerating this process, high error rates and frequent hallucinations still necessitate manual review.
We propose a fully automated approach using novel techniques to improve accuracy and mitigate hallucinations. 


\subsection{Reference-guided Extraction}
The idea of using external references or knowledge sources to guide information extraction has been explored in various domains, including clinical NLP~\cite{demner2005knowledge}. 
Researchers have investigated the use of medical ontologies, knowledge bases, and domain-specific corpora to improve the performance of clinical information extraction systems~\cite{goswami2019ontological, jin2022biomedical, kiritchenko2010exact}.
These approaches typically involve incorporating external knowledge sources into the model architecture, or using them as auxiliary inputs during training or inference. 
However, existing methods may not fully leverage the evolving knowledge available in clinical references~\cite{yan2024corrective}. 
In contrast, our system dynamically aligns and refines references with external knowledge, allowing for easy updates as new knowledge becomes available.

\section{Conclusions}
\label{sec:conclusions}
In this paper, we propose a novel framework for extracting lung lesion information from clinical and imaging reports using LLMs. Our approach aligns internal and external knowledge through in-context learning (ICL) to enhance the reliability and accuracy of extracted information. 
By dynamically selecting and updating internal knowledge and using external knowledge solely for internal-knowledge updates, our method outperforms commonly used ICL methods over data from real-world clinical trials.  
It excels in accurately detecting and extracting the most clinically relevant lesion information, such as lesion size, margin, and solidity.

\section*{Ethical Considerations}
We recognize the importance of meeting all ethical and legal standard throughout our work, particularly in handling sensitive medical data and PII.

The clinical data used in this study may not be shared or distributed. 
All PII in the data used for this work have been fully redacted, to protect patient identities and adhere strictly to all relevant regulations, laws and guidelines.
Our commitment to data security extends to our model development process, which is limited to the use of privacy friendly Google Cloud LLMs.
This tool has been approved by our Data Governance Committee, ensuring that our practices align with institutional guidelines and maintain the highest standards of data security and compliance.



\section*{Acknowledgments}
This work is supported by Freenome. 
We thank Chuanbo Xu for assistance in acquiring the imaging and clinical reports, and Nasibeh Vatankhah for providing clinical insights and helping to develop the lung lesion field schema.

\bibliography{custom}
\newpage
\appendix
\section{Appendix}
\label{sec:appendix}

\subsection{Data Preparation and Annotation}
\label{sec:appendix_data}
\begin{table}
    \centering
    \small
    \begin{tabular}{lccc}
    \toprule
    \textbf{Lesion} & \textbf{Total} & \textbf{Training} & \textbf{Test}\\
    \midrule
    Subjects & 19 & 9 & 10\\
    Clinical reports & 31 & 16 & 15\\
    Imaging reports & 30 & 14 & 16\\ 
    Total findings & 189 & 81 & 108\\
    \bottomrule
    \end{tabular}
    \caption{Manually annotated lung lesion dataset statistics.}
    \label{tab:stats}
    \end{table}
We use a real-world dataset collected from a case-control, multicenter diagnostic study designed to gather blood samples for the development of blood-based screening tests. 
In the collected clinical and imaging reports, all personally identifiable information (PII) had been previously redacted. 
The textual information within these reports is extracted using optical character recognition (OCR) via Google's Cloud Vision API~\cite{GoogleCloudVision}.

Two annotators with clinical expertise manually identify all lung lesion findings and extract relevant fields based on our annotation schema (Table \ref{tab:guidelines}). 
The inter-annotator agreement (IAA) is assessed using 10 reports reviewed by both annotators and calculated using Cohen’s $\kappa$. 
The 10 reports include 5 clinical notes and 5 radiology reports from 2 subjects. 
The average Cohen's $\kappa$ value for 9 lesion fields is 0.86.
In cases where discrepancies are found, a third clinician participates to resolve the differences and ensure consensus.
The counts of subjects, reports, and findings in the training and test splits are listed in Table~\ref{tab:stats}.

\subsection{Lung Lesion Annotation Schema}
\label{appendix:guideline}
According to the Lung-RADS guidelines \footnote{https://www.acr.org/-/media/ACR/Files/RADS/Lung-RADS/Lung-RADS-2022.pdf}, the full annotation schema is described in Table \ref{tab:guidelines}. 

\begin{table*}[ht]
\centering
\small
\begin{tabular}{l|p{10cm}}
\toprule
\textbf{Field} & \textbf{Description} \\
\midrule
Evaluator Signed On & The medical expert who signs the report, such as a physician, medical examiner, or pathologist. The expert’s signature verifies the report and confirms their agreement with the findings and opinions. \\
\midrule
Date of Report Signed & The date the medical expert signs the report. \\
\midrule
Imaging Procedure & The imaging procedure identifying the pulmonary lesion, including documentation or comparisons of previous procedures. \\
\midrule
Date of Imaging Procedure Performed & The date the imaging procedure is performed. \\
\midrule
Lesion SeqNo & An auxiliary variable to help track the number of lesions described in a report, listed in chronological order if dates are available. \\
\midrule
Number of Lesions & Indicates whether the lesions are solitary or multiple. \\
\midrule
Lesion Size (mm) & Size can be reported in diameter, area, or all three dimensions (width, height, depth). Usually measured in millimeters; convert from centimeters if necessary. \\
\midrule
SUV & The reported standard uptake value of the nodule, which may be provided even if lesion size is not mentioned. \\
\midrule
Lesion Type & Terms used in medical imaging to describe small growths in the lungs, differing mainly in size. A pulmonary nodule is a rounded opacity $\leq$ 3 cm in diameter, while a pulmonary mass is > 3 cm. A pulmonary cyst is an air- or fluid-filled sac within lung tissue. \\
\midrule
Lobe & The lobe of the lung where the nodule is located. \\
\midrule
Lesion Description & Detailed description of the pulmonary lesion. \\
\midrule
Margin & Describes the edge characteristics of the lesion, such as `spiculated', `well-defined', or `irregular'. \\
\midrule
Solidity (Morphology) & Refers to the shape and structure of the lesion, such as `ground glass', `partly-solid', or `solid'. 
\begin{itemize}
    \item For solid and part-solid nodules, the size threshold for an actionable nodule or positive screen is $\geq$ 6 mm.
    \item For nonsolid (ground-glass) nodules, the size threshold is $\geq$ 20 mm.
    \item On follow-up screening CT exams, the size cutoff is $\geq$ 4 mm for solid and part-solid nodules and/or an interval growth of $\geq$ 1.5 mm of preexisting nodule(s).
\end{itemize} \\
\midrule
Calcification & Indicates if the pulmonary nodules are calcified. \\
\midrule
Cavitation & A gas-filled space within the lung tissue. \\
\midrule
Lung RADS Score & Lung-RADS is a classification system for findings in low-dose CT (LDCT) screening exams for lung cancer. Examples include `4A', `4B', and `4X'. \\
\bottomrule
\end{tabular}
\caption{Lung lesion annotation schema.}
\label{tab:guidelines}
\end{table*}

\subsection{Prompts}
\label{appendix:prompt}
The system prompts for rule generator, grader, lung lesion finding detection, and lesion descrition extraction are presented in Table \ref{tab:rule_generator}, \ref{tab:grader_prompt}, \ref{tab:finding_detection}, and \ref{tab:char_extraction}, respectively.

\begin{table*}[ht]
\centering
\small
\begin{tabular}{l|p{10cm}}
\toprule
\textbf{Role} & \textbf{Prompt} \\
\midrule
System & \textit{You are a pulmonary radiologist. Your task is to extract key findings from the clinical or imaging reports.} \\
\midrule
User & How many findings of Lung Lesions are present in the following text: \texttt{\{text\}} \\
System & \texttt{\{lesion\_number\}} \\
User & Please provide detailed explanations. \\
System & \texttt{\{detailed\_explanations\}} \\
User &  Only \texttt{\{num\_findings\}} findings should be classified as Lung Lesions, explain why they are and why the remaining findings are not. Return in JSON format of: \texttt{\{"lung lesion findings": ["referred text": "reason of being lung lesion finding"], "none lung lesion findings": ["referred text": "reason of not being lung lesion finding"]\}} \\
System & \texttt{\{references\}} \\
User & Transform the references into generalized, reusable rules by abstracting common properties. Format the output in the following JSON structure:
\texttt{[{"pattern": "example pattern", "rule": "example rule description"}]} \\
System & \texttt{\{lung-relevant rules, lung-irrelevant rules\}} \\
\bottomrule
\end{tabular}
\caption{Multi-dialogue prompt template of rule generator.}
\label{tab:rule_generator}
\end{table*}

\begin{table*}[h]
\centering
\small
\begin{tabular}{p{0.95\textwidth}}
\toprule
\textit{You are a grader assessing the helpfulness and truthfulness of retrieved rules related to pulmonary (lung) lesions in the context of pulmonary lesion findings.}. \\
\midrule
Given the clinical or imaging report, please evaluate the \textit{helpfulness} of each rule on a scale from 1 to 5, where:
\begin{minipage}{\textwidth}
\begin{Verbatim}[fontsize=\small]
    1 means not helpful at all
    2 means slightly helpful
    3 means moderately helpful
    4 means very helpful
    5 means extremely helpful
\end{Verbatim}
\end{minipage}
\\
\midrule
Below is the clinical or imaging report: \\
\texttt{\{input\_query\}} \\
\midrule
Additionally, evaluate the \textit{truthfulness} of each rule based on the retrieved knowledge on a scale from 1 to 3, where:
\begin{minipage}{\textwidth}
\begin{Verbatim}[fontsize=\small]
    1 means not truthful at all
    2 means partially truthful
    3 means completely truthful
\end{Verbatim}
\end{minipage}
\\
Provide a brief explanation indicating how the rule can help in the extraction of pulmonary lesion characteristics and how the retrieved knowledge supports or refutes the rule. \\
\midrule
Below is the retrieved external knowledge: \\
\texttt{\{external\_knowledge\}} \\
\bottomrule
\end{tabular}
\caption{Prompt template for grader to assess helpfulness and truthfulness. Note that we chose a range score of 1-5 for truthfulness in our sample study, but extreme values of 1 and 5 are rare, so we set the range to 1-3.}
\label{tab:grader_prompt}
\end{table*}

\begin{table*}[h]
\centering
\small
\begin{tabular}{p{0.95\textwidth}}
\toprule
\textit{You are a pulmonary radiologist. Extract key findings from the clinical or imaging report and organize them into the provided JSON structure}. \\
\midrule
Use the following JSON template as a guide: \\
\begin{minipage}{\textwidth}
\begin{Verbatim}[fontsize=\small]
[
    {
        "Imaging Procedure": "Enter imaging procedure here or 'None'",
        "Procedure Date": "Enter date in YYYY-MM-DD format here or 'None'",
        "Lung RADS": "Enter Lung RADS category here or 'None'",
        "Number of Lesion": "Enter number of lesion here or 'None'",
        "Lagest Lesion Size": "Enter lesion size here",
        "Lesion Type": "Enter lesion type here",
        "SUV": "Enter SUV here or 'None'",
        "Location": "Enter location here or 'None'",
        "Lesion Description": "Enter Lesion Description here or 'None'",
        "Text Source": "Enter text source here or 'None'",
    },
    {
        // Add additional finding as needed
    
    },
]      // Lung Lesion Findings
\end{Verbatim}
\end{minipage} \\
\midrule
Below is the clinical or imaging report: \\
\texttt{\{input\_query\}} \\
\midrule
Below are some examples for reference: \\
\texttt{\{few\_shot\_samples\}} \\
\midrule
Below are some lung-related rules for reference: \\
\texttt{\{corrected\_rules\}} \\
Below are some lung-irrelevant rules for reference: \\
\texttt{\{corrected\_rules\}} \\
\bottomrule
\end{tabular}
\caption{Prompt template for stage-1 lung lesion finding detection.}
\label{tab:finding_detection}
\end{table*}

\begin{table*}[h]
\centering
\small
\begin{tabular}{p{0.95\textwidth}}
\toprule
\textit{You are a pulmonary radiologist. 
Please extract location description, margin, solidity, calcification, cavitation from lesion description and organize them into the provided JSON structure}. \\
\midrule
Use the following JSON template with \textbf{preferred vocabularies} as a guide: \\
\begin{minipage}{\textwidth}
\begin{Verbatim}[fontsize=\small]
{
    "location description": "Enter location description here or 'None'",
    "margin": "Enter margin description here, preferably from the vocabulary 
                ['spiculated', 'rounded', 'ill-defined', 'irregular', 'lobulated'] or 'None'"
    "solidity": "Enter solidity description only from the fixed vocabulary ['solid', 
                'partly solid', 'groundglass', 'ground-glass', 
                'groundglass and consolidative'] or 'None'",
    "calcification": "Enter calcification description here,     
                    preferably from ['noncalcified'] or 'None'",
    "cavitation": "Enter cavitation description here, 
                    preferably from ['mildly cavitary', 'cavitary'] or 'None'"
}
\end{Verbatim}
\end{minipage}
\\
\midrule
Below is the lesion description text: \\
\texttt{\{lesion\_description\_text\}} \\
\midrule
Below is the full text of report containing the finding for reference: \\
\texttt{\{source\_text\}} \\
\midrule
Below are some examples for reference: \\
\texttt{\{few\_shot\_samples\}} \\
\bottomrule
\end{tabular}
\caption{Prompt template for stage-2 lesion description text structured data extraction.}
\label{tab:char_extraction}
\end{table*}

\subsection{Hyper-parameters}
\label{appendix:para}
The hyper-parameter settings for all modules are listed in Table~\ref{tab:parameter}.
\begin{table*}[ht]
\centering
\small
\begin{tabular}{l|l|c}
    \toprule
    \textbf{Module} & \textbf{Hyper-parameter} & \textbf{Value} \\
    \midrule
    \multirow{2}{*}{\textbf{Rule generator}} & temperature & 0.9 \\
     & top\_p & 1 \\
    \midrule
    \multirow{3}{*}{\textbf{Retriever}} & retrival threshold for external knowledge & 0.9 \\
     & retrieved top-$k$ lung-related rule & 2 \\
     & retrieved top-$k$ lung-irrelevant rule & 1\\
    \midrule
    \multirow{3}{*}{\textbf{Grader}} & number of interations $I$ & 3 \\
     & \textit{truthfulness} threshold  & 2 \\
     & \textit{helpfulness} threshold & 4 \\
    \midrule
    \textbf{Lesion finding detection} & temperature & 0.2 \\
    \midrule
    \textbf{Lesion description extraction} & temperature & 0.2 \\
    \bottomrule
\end{tabular}
\caption{Hyper-parameter settings used by our clinical data extraction system.}
\label{tab:parameter}
\end{table*}

\end{document}